\documentclass[letterpaper, 10 pt, conference]{ieeeconf}  % Comment this line out if you need a4paper

\IEEEoverridecommandlockouts                              % This command is only needed if 
\title{\LARGE \bf
Joint Multi-Object Detection and Tracking with Camera-LiDAR Fusion for Autonomous Driving
}

\author{Kemiao Huang$^{1}$ and Qi Hao$^{1,2,3*}$% <-this % stops a space
\thanks{This work is partially supported by the Shenzhen Fundamental Research Program (No: JCYJ20200109141622964), and the Intel ICRI-IACV Research Fund (CG\#52514373).}%
\thanks{$^{*}$Corresponding author: Qi Hao (hao.q@sustech.edu.cn).}%
\thanks{$^{1}$Department of Computer Science and Engineering, Southern University of Science and Technology, Shenzhen 518055, China.}%
\thanks{$^{2}$Sifakis Research Institute of Trustworthy Autonomous Systems, Southern
University of Science and Technology, Shenzhen 518055, China.}
\thanks{$^{3}$Pazhou Lab, Guangzhou 510330, China.}
}

\usepackage{algorithm}
\usepackage{algorithmic}
\usepackage{cite}
\usepackage{amsmath}
\usepackage{bm}
\usepackage{amsfonts}
\usepackage{graphicx}
\usepackage{amssymb}
\usepackage{multirow}
\usepackage{array}
\usepackage{etoolbox}

\makeatletter
\patchcmd{\@makecaption}
  {\scshape}
  {}
  {}
  {}
\makeatletter
\patchcmd{\@makecaption}
  {\\}
  {.\ }
  {}
  {}
\makeatother

\makeatletter
\let\NAT@parse\undefined
\makeatother

\usepackage{hyperref} 
\hypersetup{
    colorlinks=true,
}

\newcolumntype{L}[1]{>{\raggedright\let\newline\\\arraybackslash\hspace{0pt}}m{#1}}
\newcolumntype{C}[1]{>{\centering\let\newline\\\arraybackslash\hspace{0pt}}m{#1}}
\newcolumntype{R}[1]{>{\raggedleft\let\newline\\\arraybackslash\hspace{0pt}}m{#1}}

\begin{document}

\maketitle
\thispagestyle{empty}
\pagestyle{empty}

\begin{abstract}
Multi-object tracking (MOT) with camera-LiDAR fusion demands accurate results of object detection, affinity computation and data association in real time. This paper presents an efficient multi-modal MOT framework with online joint detection and tracking schemes and robust data association for autonomous driving applications. The novelty of this work includes: (1) development of an end-to-end deep neural network for joint object detection and correlation using 2D and 3D measurements; (2) development of a robust affinity computation module to compute occlusion-aware appearance and motion affinities in 3D space; (3) development of a comprehensive data association module for joint optimization among detection confidences, affinities and start-end probabilities. The experiment results on the KITTI tracking benchmark demonstrate the superior performance of the proposed method in terms of both tracking accuracy and processing speed.
\end{abstract}

%%%%%%%%%%%%%%%%%%%%%%%%%%%%%%%%%%%%%%%%%%%%%%%%%%%%%%%%%%%%%%%%%%%%%%%%%%%%%%%%
\section{INTRODUCTION}
Multi-object tracking (MOT) is a central task for autonomous driving (AD) in dynamic environment perception and dataset annotation~\cite{geiger2012we,li2020sustech}. In many AD-related object detection and tracking schemes, the camera-LiDAR fusion strategy is preferred~\cite{qi2018frustum,huang2020epnet,zhang2019robust}, as the former provides high-resolution 2D information and the latter yields high-accuracy 3D measurements. Usually, the performance of sensor fusion relies on the quality of sensor calibration. Compared with single-object tracking, MOT suffers more from target occlusions especially when the number of targets is large. A typical MOT system consists of (1) sensor calibration, (2) object detection, (3) object correlation, (4) data association, and (5) track management, as shown in Fig. \ref{fig:intro}.

Despite many efforts in developing camera-LiDAR fusion based MOT systems, there are three main technical challenges to overcome:

\begin{figure}[t]
\centering
\includegraphics[width=\linewidth]{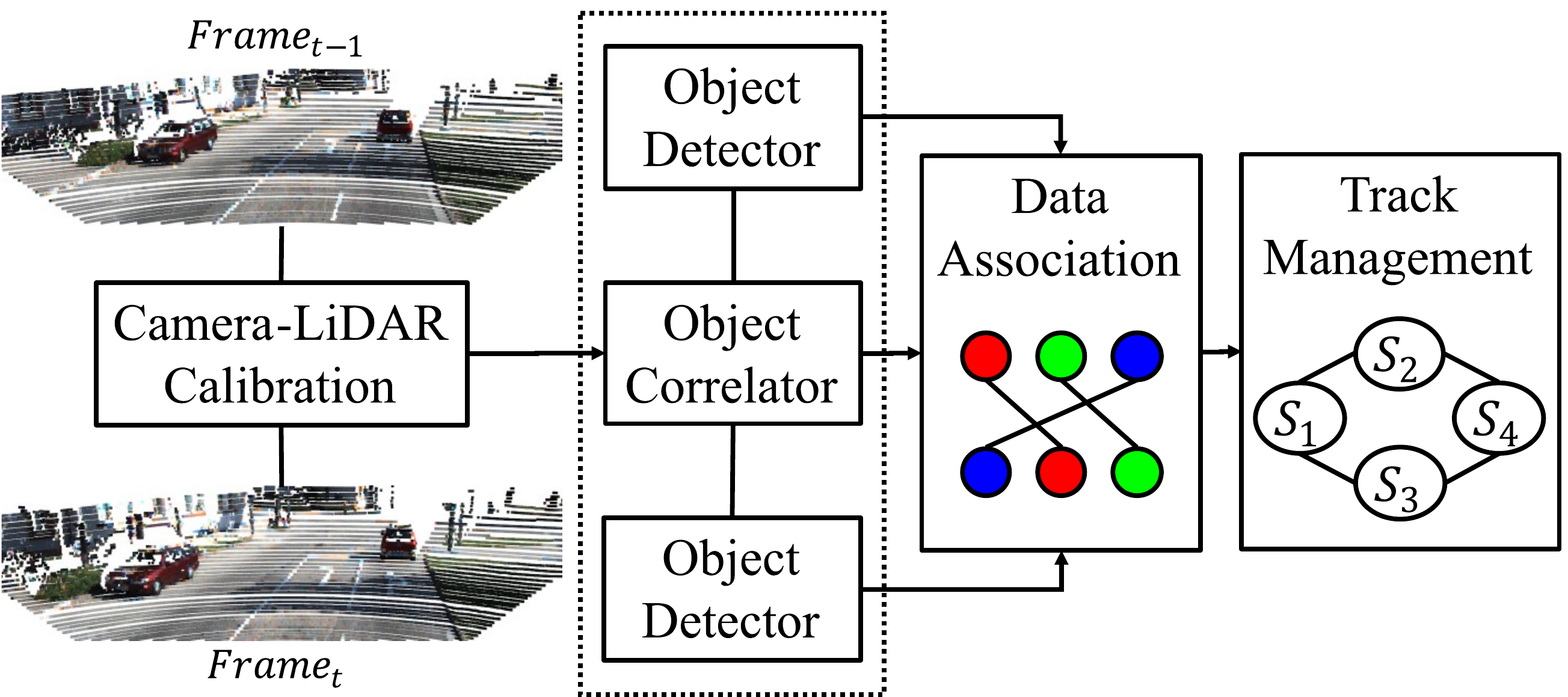}
\caption{An overview of the classical MOT system with camera-LiDAR fusion for autonomous driving. $\text{Frame}_{t-1}$ and $\text{Frame}_{t}$ are two adjacent frames of fused 2D-3D data respectively. $S_{1...4}$ represent tracking states.}
\label{fig:intro}
\end{figure}

\begin{enumerate}
    \item \textbf{Joint detection and tracking.} Most MOT methods follow the tracking-by-detection paradigm, which performs object detection, object correlation and data association in a cascade pipeline, yielding redundant computations and sub-optimal solutions. A unified framework for joint detection and tracking with parallel object detection and correlation is more suitable for real time applications.
    \item \textbf{Robust affinity computation.} The quality of object features and affinity metrics greatly affect data association performance. A robust affinity metric should incorporate appearance and motion cues to tackle the problems of tiny inter-object appearance differences, complex motions and dense distributions.
    \item \textbf{Comprehensive data association.} There are several uncertainties (\emph{e.g.} detection mistakes, occlusions, re-ID errors, start or end) that could affect tracking results. Thus, it is necessary to take these factors comprehensively into account for data association.
\end{enumerate}

Current MOT methods can fall into two groups: (1) tracking by detection (TBD)~\cite{wojke2017simple,zhang2019robust,shenoi2020jrmot}, and (2) joint detection and tracking (JDT)~\cite{wang2019towards,lu2020retinatrack,zhan2020simple,peng2020chained}.  The former trains object detection and re-identification (Re-ID) models in two isolated systems and infers tracks in a cascade manner. Such a framework likely leads to local optima and costs extra GPU computations and memories. The latter shares object features between the two parallel models but needs to train a larger network with fewer training samples~\cite{zhan2020simple}.  Most affinity estimation methods~\cite{wojke2017simple,wang2019towards,lu2020retinatrack,mykheievskyi2020learning} are based on (1) Re-ID features and (2) motion predictions to improve tracking consistency. However, 2D appearance and motion cues suffer heavily from target occlusion and overlapping due to the lack of depth information, especially for dense target distributions. Most data association methods~\cite{wojke2017simple,wang2019towards,zhan2020simple,zhang2019robust} only consider the estimated affinities to perform maximum bipartite matching~\cite{kuhn1955hungarian}, while classification confidences from detection models are only used to reduce false positives in pre-processing steps.

\begin{table*}[t]
\caption{A methodological comparison between state-of-the-art MOT methods and the proposed method (JMODT).}
\label{tab:related}
\centering
\begin{tabular}{c|c|c|c|C{2.65cm}|c|c|C{1.7cm}}
\hline\hline
\multirow{2}{*}{\textbf{Type}} &
  \multirow{2}{*}{\textbf{Method}} &
  \multicolumn{2}{c|}{\textbf{Object Detection and Correlation}} &
  \multicolumn{3}{c|}{\textbf{Affinity Metric}} &
  \multirow{2}{1.7cm}{\centering \textbf{Data Association}} \\ \cline{3-7}
 &
   &
  \textbf{Detection} &
  \textbf{Correlation} &
  \textbf{Appearance Modality} &
  \textbf{Motion} &
  \textbf{Geometry} &
   \\ \hline
\multirow{4}{1.2cm}{\centering Tracking by Detection} &
  ODESA~\cite{mykheievskyi2020learning} &
  2D &
  Re-ID &
  Camera &
  KF &
  2D Distance &
  HA \\ \cline{2-8} 
 &
  SMAT~\cite{gonzalez2020smat} &
  2D &
  Re-ID + Optical Flow &
  Camera &
  $\times$ &
  2D IoU &
  HA \\ \cline{2-8} 
 &
  JRMOT~\cite{shenoi2020jrmot} &
  3D &
  Re-ID &
  Camera + LiDAR (Batch Fusion) &
  KF &
  3D IoU &
  JPDA \\ \cline{2-8} 
 &
  mmMOT~\cite{zhang2019robust} &
  3D &
  Re-ID + Start-End &
  Camera + LiDAR (Batch Fusion) &
  $\times$ &
  $\times$ &
  MIP \\ \hline
\multirow{5}{1.2cm}{\centering Joint Detection and Tracking} &
  CenterTrack~\cite{zhou2020tracking} &
  2D / 3D &
  Paired Detection &
  Camera &
  Offset &
  2D Distance &
  Greedy \\ \cline{2-8} 
 &
  ChainedTrack~\cite{peng2020chained} &
  2D &
  Parallel Re-ID &
  Camera &
  $\times$ &
  2D IoU &
  HA \\ \cline{2-8} 
 &
  JDE~\cite{wang2019towards} &
  2D &
  Parallel Re-ID &
  Camera &
  KF &
  2D Distance &
  HA \\ \cline{2-8} 
 &
  RetinaTrack~\cite{lu2020retinatrack} &
  2D &
  Parallel Re-ID &
  Camera &
  KF &
  2D IoU &
  HA \\ \cline{2-8} 
 &
  JMODT (ours) &
  3D &
  Parallel Re-ID + Start-End &
  Camera + LiDAR (Point-Wise Fusion) &
  KF &
  3D DIoU &
  Improved MIP \\ \hline\hline
\end{tabular}

\vspace{2mm}
``KF'' means Kalman filter, ``Offset'' means the image-based deep offset prediction, ``IoU'' means intersection-over-union, ``DIoU'' means distance-IoU affinity, ``HA'' means the Hungarian algorithm, ``JPDA'' means joint probabilistic data association, ``MIP'' means mixed-integer programming.
\end{table*}

In this paper, we propose a real-time and robust Joint Multi-Object Detection and Tracking (JMODT) system that performs joint learning of 3D object detection and tracking, with robust affinity computation and comprehensive data association. The main contributions of this work include:
\begin{enumerate}
    \item Developing an end-to-end network that simultaneously generates 3D bounding boxes and association scores from camera and LiDAR measurements for real time joint detection and tracking.
    \item Developing a robust affinity computation module that combines multi-modal features and 3D motion predictions with robust affinity metrics.
    \item Developing a comprehensive data association module that takes both detection uncertainties and object correlation confidences into account.
    \item Performing experiments on the KITTI tracking benchmark~\cite{geiger2012we}. Our method outperforms the baselines in terms of both tracking accuracy and processing speed. The open-source code is available at \href{https://github.com/Kemo-Huang/JMODT}{https://github.com/Kemo-Huang/JMODT}
\end{enumerate}

The rest of this paper is organized as follows. Section \ref{sec:related} introduces the related work on MOT methods. Section \ref{sec:system} describes the system setup and problem statement. Section \ref{sec:method} presents the proposed method. Section \ref{sec:experiment} provides the experiment results and discussions. Section \ref{sec:conclusion} concludes this paper.

\section{RELATED WORK}\label{sec:related}
TABLE \ref{tab:related} summarizes most state-of-the-art MOT methods for AD applications.  For joint camera-LiDAR sensing modalities, there are only a few TBD methods~\cite{zhang2019robust,shenoi2020jrmot}, however no JDT frameworks, have been developed. LiDAR data can provide extra 3D information but limits the frame rate at the same time. Usually, JDT frameworks use the same feature and paired frames for detection and Re-ID \cite{wang2019towards,lu2020retinatrack,zhou2020tracking,peng2020chained}, thus the extra 3D information can help improve the quality of parallel Re-ID. Besides, start-end estimation networks are used to create or delete new or obsolete IDs respectively within successive frames~\cite{zhang2019robust}. In this work, we develop a two-stage detector with two parallel branches of Re-ID and start-end estimation for paired-frame inputs.

Affinities among objects can be estimated using features, motion prediction and geometric intersections. Different from perspective-based data fusion for object detection~\cite{qi2018frustum}, object-level batch feature fusion schemes have been developed for MOT, where image features and LiDAR features are concatenated to represent the multi-modal features~\cite{shenoi2020jrmot}, or in terms of attention maps~\cite{zhang2019robust}. However, those batch fusion methods extract object features separately from each modality, and do not leverage point-wise correspondences between two sets of data. In this work, we use a multi-scale point-wise feature fusion scheme~\cite{huang2020epnet} to estimate appearance affinities. Kalman filters (KFs) have been extensively used for motion predictions in MOT~\cite{thrun2002probabilistic,wojke2017simple,wang2019towards,lu2020retinatrack,zhan2020simple,mykheievskyi2020learning,shenoi2020jrmot}. Intersection-over-union (IoU) between each pair of object boxes, along with motion predictions, have been used as motion affinities~\cite{gonzalez2020smat,lu2020retinatrack,peng2020chained,shenoi2020jrmot}. However, IoU-based affinities are too strict when targets have complex motions~\cite{wojke2017simple}; on the other hand, distance-IoUs~\cite{zheng2020distance} can tolerate more challenging situations. Therefore, we develop a 3D distance-IoU based metric for motion affinity refinement.

Usually, data association is formulated as a bipartite graph matching problem and is solved using the Hungarian algorithm (HA)~\cite{kuhn1955hungarian}. Besides, a variety of Bayesian methods~\cite{shenoi2020jrmot,scheidegger2018mono} have also been developed to utilize data temporal correlation. Recently, joint optimization among classification confidences, affinities and start-end probabilities have been developed by using mixed-integer programming (MIP) and deep structure models (DSMs)~\cite{zhang2019robust,frossard2018end}. However, these methods use an extra DSM network to infer classification confidences with a redundant feature extraction process. Therefore, we propose a combined objective function to optimize both detection and tracking confidences through a unified network.

\begin{figure*}[t]
\centering
\includegraphics[width=\linewidth]{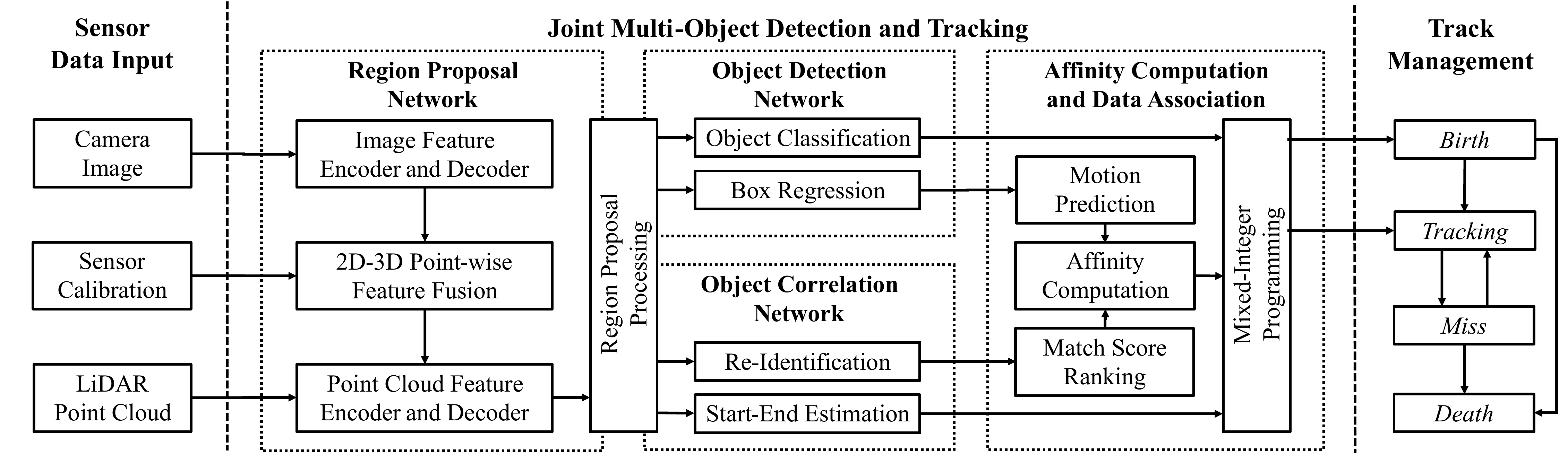}
\caption{The system architecture of the proposed camera-LiDAR based joint multi-object detection and tracking system.}
\label{fig:system}
\end{figure*}

\section{SYSTEM SETUP AND PROBLEM STATEMENT}\label{sec:system}
\subsection{System Architecture}
The architecture of our system includes five main modules: (1) region proposal network (RPN), (2) parallel detection and correlation networks, (3) affinity computation, (4) data association, and (5) track management, as shown in Fig. \ref{fig:system}. The tracking pipeline consists of five stages: (1) RPN takes calibrated sensor data from paired frames as input and generates regions of interest (RoI) and multi-modal features of the region proposals; (2) the parallel detection and correlation networks use the RoI and proposal features to generate detection results, Re-ID affinities and start-end probabilities; (3) the Re-ID affinities are further refined via the motion prediction and match score ranking modules; (4) the mixed-integer programming module performs comprehensive data association based on the detection results and computed affinities; (5) the association results are further managed to achieve continuous tracks despite object occlusions and re-appearances.

\subsection{Problem Statement}
In this work, we focus on solving the following problems:
\begin{itemize}
	\item How to properly train the parallel detection and correlation networks with shared proposal features?
	\item How to compute robust object affinities based on motion prediction results and Re-ID scores?
	\item How to achieve comprehensive data association using classification confidences, object affinities and start-end probabilities via MIP?
\end{itemize}

\section{PROPOSED METHODS}\label{sec:method}
\subsection{Parallel Object Detection and Correlation}
\begin{figure}[htb]
\centering
\includegraphics[width=\linewidth]{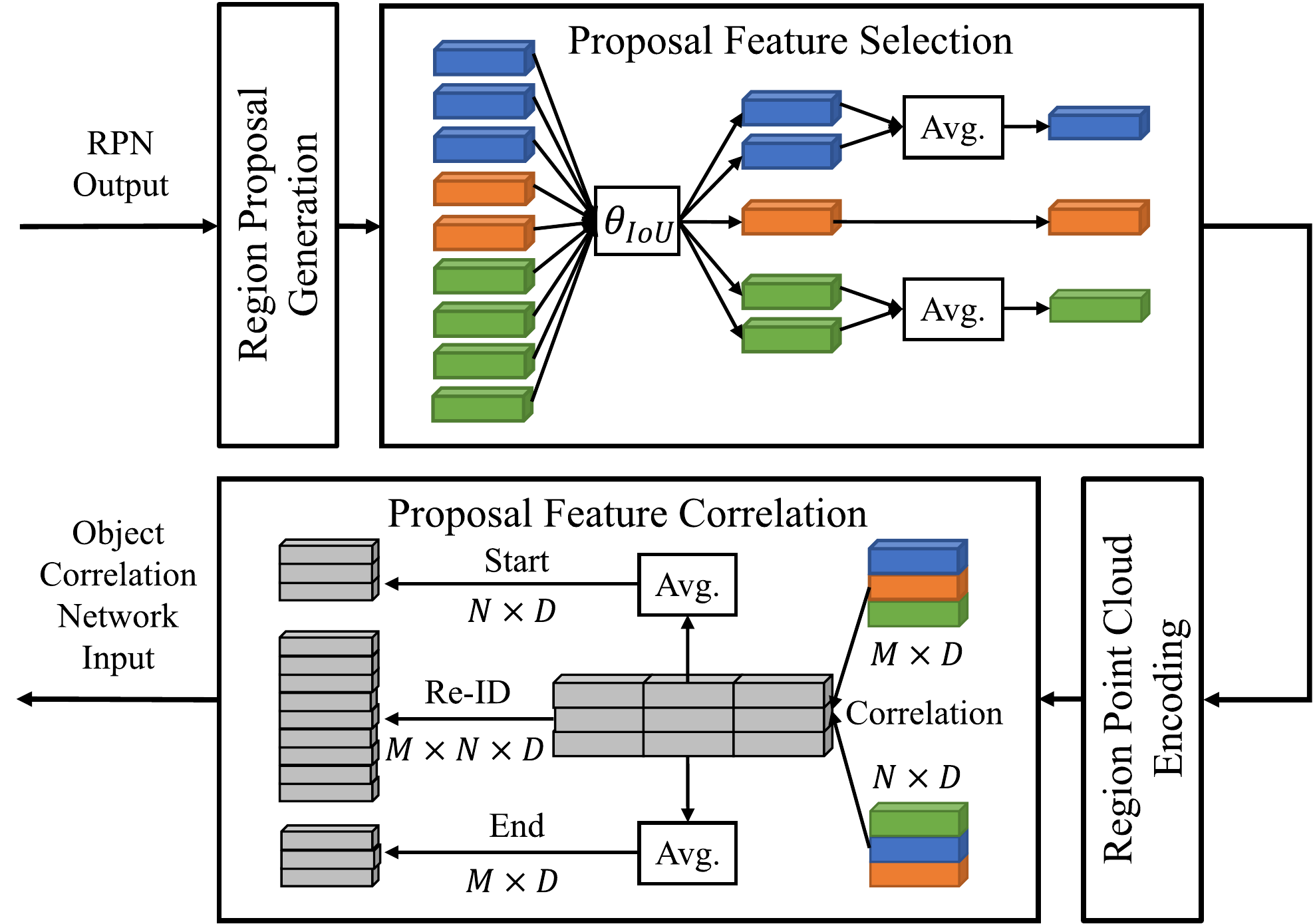}
\caption{Region proposal processing for training the object correlation network. The input proposal features with the same ID label are shown in the same color.}
\label{fig:proposal}
\end{figure}

To achieve parallelism of object detection and object correlation, the shared features of region proposals need additional processing. Without changing the region proposal generation and region point cloud encoding modules for object detection, we add two more feature processing modules (region proposal feature selection and correlation) for object correlation, as shown in Fig. \ref{fig:proposal}. Besides, the generated proposal features are saved in memory to avoid duplicate computation during network inference.

\subsubsection{Proposal Feature Selection}
The foreground region proposals generated by RPN usually have low overlaps (\emph{e.g.} $60\%$) with groundtruth boxes because they are originally used to train object detection network. Besides, those coarse proposals cannot be directly used in recognition tasks due to incomplete object information and object overlapping issues. Thus, we propose two operations to help proposal-region alignment during the correlation network training: 
\begin{itemize}
    \item Set a high IoU threshold $\theta_{iou}$ for the proposal input.
    \item Compute the average of encoded proposal features which belong to the same target ID. 
\end{itemize}
Generally, the former operation filters out the useless inputs to guarantee the convergence of training. The latter operation improves the proposal features by using the shared information and providing the missing information. On the other hand, a high classification confidence threshold $\theta_{cls}$ should also be set for the network inference because there is no supervision on background proposals in the feature correlation network after the proposal feature selection. In this work, we set $\theta_{iou}=\theta_{cls}=0.85$ empirically.

\subsubsection{Proposal Feature Correlation}
Before learning Re-ID and start-end confidences, we adopt absolute subtraction~\cite{zhang2019robust} as the pair-wise correlation operation for the proposal features to represent the target dependency between adjacent frames. Given $M$ selected proposals in frame $t-1$ and $N$ selected proposals in frame $t$, the correlated feature matrix then has the size of $M \times N$. To obtain global inter-object information, the feature matrix is averaged along its rows and columns separately. Since start-end estimation is a symmetric task, the generated $N$ ``start'' features and $M$ ``end'' features are batched to feed one independent start-end network. 

\subsubsection{Object Correlation Network Specifics and Loss Function}
We use fully connected layers for both the Re-ID and start-end estimation networks. Each selected proposal feature can be assigned to a unique ID label. Each pair of selected features should have a binary label for ID matching. Thus, both Re-ID and start-end estimation become a sort of binary classification tasks. We use softmax ranking~\cite{zhang2019robust} for Re-ID outputs and sigmoid activation for start-end outputs to map all confidences to $[0,1]$. We use L1 loss for the network training.

\subsection{Affinity Computation}
The affinity between each pair of objects should take into account both appearance and motion information. Appearance affinities are the softmax ranking results of Re-ID network outputs. Motion affinities are computed based on the geometric similarities between detected object boxes and predicted object boxes.
The pseudocode of affinity computation is shown in \textbf{Algorithm \ref{alg:affinity}}.
\subsubsection{Motion Prediction and Update}
In this work, Kalman filter (KF)~\cite{thrun2002probabilistic} is used for motion prediction. The motion state $s^{kf}$ of each object is represented as:
\begin{equation}
s^{kf} = (x,y,z,l,w,h,a,v_x,v_y,v_z)
\end{equation}
where $(x,y,z)$ denotes the box center location, $(l,w,h)$ denotes the box size, $a$ denotes the box heading angle, $(v_x,v_y,v_z)$ denotes the location change of box centers across consecutive frames (\emph{i.e.}, the linear velocity). The prediction equation of $s^{kf}$ is defined as: 
\begin{equation}\label{eq:kf_predict}
\hat{\mu}_{t+1}=A\mu_t
\end{equation}
where $\hat{\mu}_{t+1}$ is the predicted mean of $s^{kf}$ at time $t+1$, $\mu_t$ is the estimated mean of $s^{kf}$ at time $t$ and $A$ is the state transition matrix based on the constant velocity assumption. The state update equation is defined as:
\begin{equation}\label{eq:kf_update}
\mu_{t+1} = \hat{\mu}_{t+1} + K_{t+1}(o_{t+1}-H\hat{\mu}_{t+1})
\end{equation}
where $o_{t}$ is the observation at time $t$, $H$ is the measurement matrix, the Kalman gain $K_{t+1}$ at time $t+1$ is defined as:
\begin{equation}
K_{t+1} = P_{t+1}H^T(HP_{k+1}H^T+R)^{-1}
\end{equation}
where $P_{t+1}$ is the state covariance at time $t+1$ and $R$ is the measurement covariance. The update equation of $P$ is defined as:
\begin{equation}
P_{t+1} = AP_tA^T + Q
\end{equation}
where $Q$ is the process covariance. In this work, $P_0$, $H$ and $R$ are simply estimated by the detection results. The predicted mean $\hat{\mu}_{t+1}$ of $s^{kf}$ is directly used for the 3D bounding box prediction $B_{t+1}$ at time $t+1$.
\subsubsection{Affinity Metrics}
Inspired by Distance-IoU loss~\cite{zheng2020distance} in object detection, we propose 3D-DIoU affinity $\mathbf{A}^{diou}=\{a^{diou}_{d,k}, d\in\mathbf{D}, k\in\bm{K}\}$ for motion-based association between detection measurements $\bm{D}$ and tracks $\bm{K}$:
\begin{equation}
a^{diou}_{d,k} = (1 - \frac{\rho(b_d,b_k)}{l}) + \frac{B_d \cap B_k}{B_d \cup B_k}
\end{equation}
where $b_i$ denotes the center of the 3D bounding box $B_i$ for object $i$, $\rho$ is the Euclidean distance and $l$ is the diagonal length of the smallest box covering the two boxes. The former $DIS=1 - \frac{\rho(b_d,b_k)}{l}$ term helps to complement the latter $IOU=\frac{B_d \cap B_k}{B_d \cup B_k}$ term when the predicted box does not overlap with any other detected box. The refined affinity $\bm{X}^{aff}$ is the weighted sum of the appearance affinity $\bm{A}^{app}$ and the motion affinity $\bm{A}^{diou}$:
\begin{equation}\label{eq:affinity}
\bm{X}^{aff} = \alpha \bm{A}^{app} + \beta \bm{A}^{diou}
\end{equation}
where $\alpha + \beta = 1$.

\begin{algorithm}
    \caption{Affinity Computation}
    \label{alg:affinity}
    \begin{algorithmic}[1]
        \REQUIRE detection measurements $\bm{D}$, tracks $\bm{K}$ and their proposal features $\bm{F} = \{F_i, i\in \bm{D}\cap\bm{K}\}$.
        \ENSURE refined affinities $\bm{X}^{aff}=\{x^{aff}_{d,k}, d\in \bm{D}, k \in \bm{K}\}$.
        \FOR{each $k\in \bm{K}$}
        \STATE $B_k \gets$ 3D box prediction for track $k$ using KF.
        \FOR{each $d\in \bm{D}$}
        \STATE $F_{d,k} \gets$ Feature correlation $|F_d - F_k|$
        \STATE $a^{app}_{d,k} \gets$ Appearance Re-ID for feature $F_{d,k}$.
        \STATE $B_d \gets$ 3D box generation for detection $d$.
        \STATE $a^{diou}_{d,k} \gets \left(1 - \frac{\rho(b_d,b_t)}{c}\right) + \frac{B_d \cap B_t}{B_d \cup B_t}$
        \ENDFOR
        \ENDFOR
        \STATE $\bm{A}^{app}\gets \{a^{app}_{d,k}, d\in \bm{D}, k \in \bm{K}\}$
        \STATE $\bm{A}^{diou}\gets \{a^{diou}_{d,k}, d\in \bm{D}, k \in \bm{K}\}$
        \STATE $\bm{P} \gets$ Softmax $\bm{A}^{app}$ along columns.
        \STATE $\bm{Q} \gets$ Softmax $\bm{A}^{app}$ along rows.
        \STATE $\bm{A}^{app} \gets \frac{1}{2} (\bm{P}+\bm{Q})$
        \STATE $\bm{X}^{aff} \gets \alpha \bm{A}^{app} + \beta \bm{A}^{diou}$
    \end{algorithmic}
\end{algorithm}

\subsection{Mixed-Integer Programming for Data Association}
In the formulation of mixed-integer programming for data association, target matching states are represented by three types of binary integer variables:
\begin{equation}
    \bm{Y} = \left[y^{cls}, y^{aff}, y^{se}\right]
\end{equation}
where $y^{cls}$ denotes whether an object is true positive, $y^{aff}$ denotes whether a measurement and a track are matched as the same object, $y^{se}$ denotes whether a measurement starts a new ID or a track ends its obsolete ID (\emph{i.e.}, an object which should not be matched by others). The corresponding three types of linear confidences are:
\begin{equation}
    \bm{X} = [x^{cls}, x^{aff}, x^{se}]
\end{equation}
where $x^{cls}$ denotes the classification confidences from the object detection network, $x^{aff}$ denotes the refined object affinities from the affinity computation module and $x^{se}$ denotes start-end probabilities from the start-end estimation network. Given detection measurements $\bm{D}$ and tracks $\bm{K}$, the constraints for association variables are straightforward:
\begin{equation}
    \forall d \in \bm{D}, y_d^{cls} = \sum_{k} y_{d,k}^{aff} + y_d^{se}
\label{equ:constraint1}
\end{equation}
\begin{equation}
    \forall k \in \bm{K}, y_k^{cls} = \sum_{d} y_{d,k}^{aff} + y_k^{se}
\label{equ:constraint2}
\end{equation}
We propose the objective function of MIP as: 
\begin{equation}\label{eq:objective}
    \mathop{\arg\max}\limits_{y} [w^{cls}(x^{cls} - 1), w^{aff}x^{aff}, w^{se}x^{se}]\bm{Y}^T
\end{equation}
where $w^{cls}$, $w^{aff}$ and $w^{se}$ are positive weights.
We modify the positive coefficient $x^{cls}$ in previous work~\cite{frossard2018end,zhang2019robust} to a negative coefficient $(x^{cls} - 1)$ because the detection score should be a penalty term to prevent matching or creating trajectories for false positive measurements (\emph{i.e.}, if the classification confidence of a target is low, then $y^{cls}$ for that target is more likely to be zero). Besides, additional weights are needed for the three terms because there is no constraint between the linear confidences during the network training. These weights help to generate longer trajectories and fewer false outputs. The pipeline of MIP-based data association is shown in \textbf{Algorithm \ref{alg:mip}}.

\begin{algorithm}
    \caption{MIP-based Data Association}
    \label{alg:mip}
    \begin{algorithmic}[1]
        \REQUIRE detection measurements $\bm{D}$, tracks $\bm{K}$, classification confidences $\bm{X}^{cls}$, object affinities $\bm{X}^{aff}$ and start-end probabilities $\bm{X}^{se}$. 
        \ENSURE association results $\bm{Y} = [y^{cls}, y^{aff}, y^{se}]$.
		\STATE Initialize binary integer variables $y^{cls}, y^{aff}$ and $y^{se}$.
        \FOR{each $d\in D$}
        \STATE $c^{cls}_d \gets w^{cls}(x^{cls}_d - 1)$
        \STATE $c^{se}_d \gets w^{se}x^{se}_d$
        \STATE Add constraints: $y_d^{cls} = \sum_k y_{d,k}^{aff} + y_d^{se}$
        \ENDFOR
        \FOR{each $k\in \bm{K}$}
        \STATE $c^{cls}_k \gets w^{cls}(x^{cls}_k - 1)$
        \STATE $c^{se}_k \gets w^{se}x^{se}_k$
        \STATE Add constraints: $y_k^{cls} = \sum_d y_{d,k}^{aff} + y_k^{se}$        
        \ENDFOR
        \FOR{each $d\in D$}
        \FOR{each $t\in T$}
        \STATE $c^{aff}_{d,k} \gets w^{aff}x^{aff}_{d,k}$
		\ENDFOR        
		\ENDFOR        
		\STATE $\bm{C} \gets [c^{cls}, c^{aff}, c^{se}]$
        \STATE Compute MIP solution: $\bm{Y} \gets \mathop{\arg\max}\limits_{y} \bm{C}\bm{Y}^T$
    \end{algorithmic}
\end{algorithm}

\section{EXPERIMENT RESULTS}\label{sec:experiment}
This work uses the KITTI tracking benchmark~\cite{geiger2012we} as the evaluation platform. The dataset provides 21 training sequences and 29 test sequences of front-view camera images and LiDAR point clouds. The training sequences are split into a training set and a validation set with roughly equal number of frames. Specifically, the training set has 10 sequences and 3975 frames, and the validation set contains 11 sequences and 3945 frames. Each ground truth (GT) instance in a frame contains a 3D bounding box and with an unique ID. Only the track has 2D IoU with a GT box greater than 0.5 can be accepted as a true positive (TP). Tracks without matched GT boxes are regarded as false positives (FPs). GT boxes without matched tracks are regarded as false negatives (FNs). Following the KITTI benchmark, we use CLEARMOT, MT/PT/ML, ID switches (IDSW), fragmentations (FRAG) and runtime per frame~\cite{bernardin2008evaluating,li2009learning} to evaluate MOT performance.

Our system is implemented with PyTorch~\cite{paszke2017automatic}. The pre-trained detection model of EPNet~\cite{huang2020epnet} is used. The correlation network is trained for 50 epochs with batch size 12. We use AdamW~\cite{loshchilov2017decoupled} optimizer with cosine annealing learning rate~\cite{loshchilov2016sgdr} 2e-4. For affinity computation, we set $\beta = 10\alpha$. For data association, we set $w^{cls}=100$, $w^{aff}=22$ and $w^{se}=1$. The hyperparameters are tuned via cross validation. For track management, we discard the traditional track incubation process because detection uncertainties have been considered in MIP. We set hit threshold $\theta_{hit}=0$ and miss threshold $\theta_{miss}=2$. Further, we keep the tracks whose association results $y^{cls}=0$ as the tentative tracks and set their initial miss times $n_{miss}=1$.

We remove the appearance affinity, IoU affinity and distance affinity alternately to perform the ablation study, as shown in TABLE \ref{tab:affinity}. It can be seen that the use of all three affinities together can greatly improve the tracking performance.

\begin{table}[ht]
\caption{Evaluation of different metrics for affinity computation.}
\label{tab:affinity}
\centering
\begin{tabular}{c|c c c c}
\hline\hline
Affinity    & FP$\downarrow$  & FN$\downarrow$   & IDSW$\downarrow$ & MOTA$\uparrow$    \\ \hline
APP         & 520 & 1098 & 659 & 79.51\% \\
DIS         & 637 & 954  & 5   & 85.64\% \\
IOU         & 532 & 1067 & 52  & 85.14\% \\
APP+DIS     & 534 & 1037 & 4   & 85.82\% \\
APP+IOU     & 556 & 1033 & 6   & 85.65\% \\
DIS+IOU     & 588 & 987  & 4   & 85.79\% \\
APP+DIS+IOU & 578 & 975  & 2   & 86.01\% \\ \hline\hline
\end{tabular}
\end{table}

\begin{figure*}[ht]
\centering
\includegraphics[width=\linewidth]{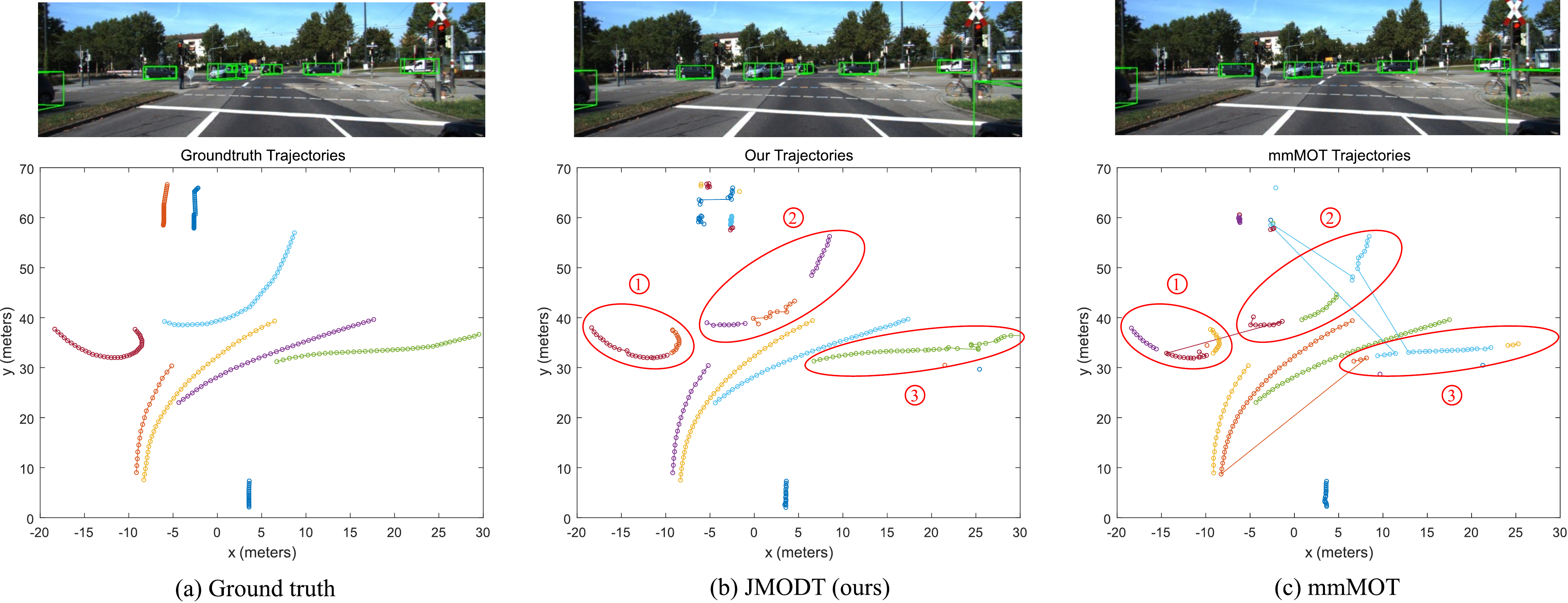}
\caption{A comparison of bird's eye view trajectories between our method and mmMOT using a KITTI data sequence. Trajectories with the same ID are shown in the same color. Region 1 indicates the case of complex object motion. Region 2 and 3 indicate the cases of object occlusions.}
\label{fig:crowd}
\end{figure*}

To evaluate the impact of measurement uncertainties upon data association, we use the Hungarian algorithm (HA) as the baseline, where HA-based data association assumes all inputs are true positives. TABLE \ref{tab:measurement} shows the  tracking performance using three sets of different qualities of inputs for two data association schemes. It can be seen that the proposed MIP method performs much better than HA.

\begin{table}[ht]
\caption{Evaluation of HA and MIP for data association.}
\label{tab:measurement}
\centering
\begin{tabular}{c|c|c c c c}
\hline\hline
$\theta_{cls}$        & Association  & FP$\downarrow$  & FN$\downarrow$   & IDSW$\downarrow$ & MOTA$\uparrow$    \\ \hline
\multirow{2}{*}{0.85} & HA  & 1946 & 834  & 1169 & 64.47\% \\  
                      & MIP & 578  & 975  & 2    & 86.01\% \\ \hline
\multirow{2}{*}{0.90} & HA  & 1781 & 881  & 1170 & 65.52\% \\  
                      & MIP & 561  & 1000 & 1    & 85.94\% \\ \hline
\multirow{2}{*}{0.95} & HA  & 1466 & 1008 & 1087 & 67.96\% \\ 
                      & MIP & 483  & 1101 & 3    & 85.72\% \\ \hline\hline
\end{tabular}
\end{table}

\begin{table*}[t]
\caption{A comparison of tracking performance of camera-LiDAR based MOT methods on the test set of the KITTI car tracking benchmark.}
\label{tab:test_results}
\centering
\begin{tabular}{c|c c c c c c c c c c c}
\hline\hline
Method &
  %Sensers &
  JDT &
  AT &
  \textcolor{red}{\textbf{MOTA$\uparrow$}} &
  MOTP$\uparrow$ &
  FP$\downarrow$ &
  FN$\downarrow$ &
  MT$\uparrow$ &
  ML$\downarrow$ &
  IDSW$\downarrow$ &
  FRAG$\downarrow$ &
  \textcolor{red}{\textbf{Runtime$\downarrow$}} \\ \hline
JRMOT~\cite{shenoi2020jrmot} &
  %Camera, LiDAR &
  $\times$ &  
  $\checkmark$ &
  85.70\% &
  \textbf{85.48\%} &
  772 &
  4049 &
  71.85\% &
  4.00\% &
  98 &
  \textbf{372} &
  0.07s \\
mmMOT~\cite{zhang2019robust} &
  %Camera, LiDAR &
  $\times$ &  
  $\checkmark$ &
  84.77\% &
  85.21\% &
  \textbf{711} &
  4243 &
  73.23\% &
  \textbf{2.77\%} &
  284 &
  753 &
  0.02s \\
JMODT (ours) &
  %Camera, LiDAR &
  $\checkmark$ &
  $\times$ &
  \textbf{86.27\%} &
  85.41\% &
  1244 &
  \textbf{3433} &
  \textbf{77.38\%} &
  2.92\% &
  \textbf{45} &
  586 &
  \textbf{0.01s} \\ \hline\hline
\end{tabular}

\vspace{2mm}
``JDT'' means joint detection and tracking. ``AT'' means the use of additional data sources for training. ``MOTA'' means MOT accuracy. ``MOTP'' means MOT precision. ``MT'' means mostly tracked. ``ML'' means mostly lost. The data come from \href{http://www.cvlibs.net/datasets/kitti/old\_eval\_tracking.php}{http://www.cvlibs.net/datasets/kitti/old\_eval\_tracking.php}.
\end{table*}

\begin{figure}[ht]
\centering
\includegraphics[width=0.95\linewidth]{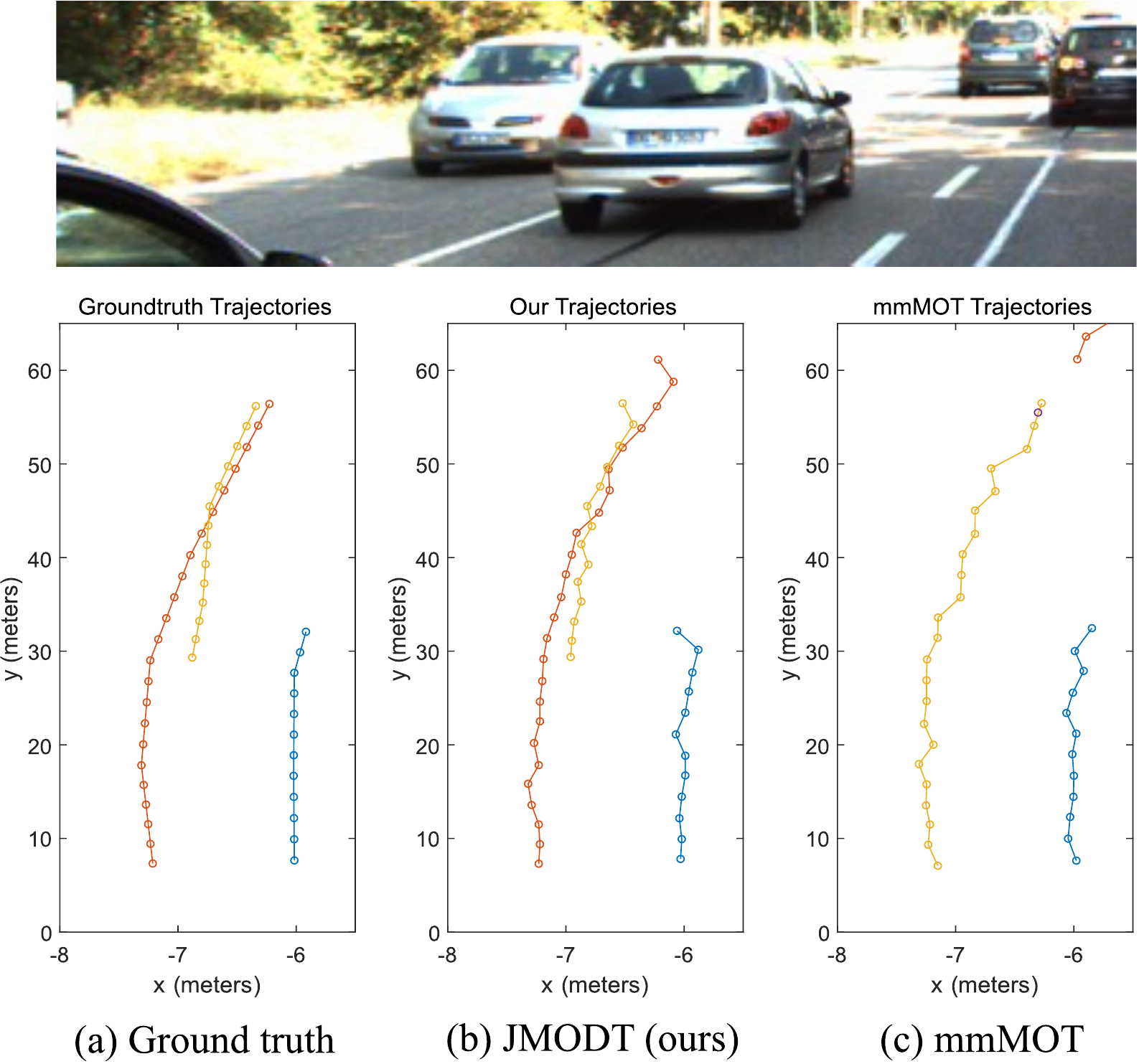}
\caption{A comparison of tracking trajectories between our method and mmMOT in a KITTI data sequence where two objects have the similar appearance.}
\label{fig:similar}
\end{figure}

We choose the open-source camera-LiDAR fusion based method mmMOT~\cite{zhang2019robust} as our baseline to perform the case study. We use two cases to illustrate the superior performance of the developed MOT method. Case 1: the proposed robust object affinity computation module helps to produce long trajectories with fewer ID switches, despite complex object motions and object occlusions in crowded scenes, as shown in Fig. \ref{fig:crowd}. Case 2: the proposed start-end estimation module helps to create a new track even though two objects have the similar appearance, whereas the baseline method fails, as shown in Fig. \ref{fig:similar}.

The results of this work (submitted to the KITTI test server) and other published methods on the test set of the KITTI car tracking benchmark are shown in TABLE \ref{tab:test_results}. We follow the routine to report the running time of the tracking stage for a fair comparison. Our method outperforms all the reported camera-LiDAR fusion based methods in terms of MOTA and running speed. Most published methods pre-trained their networks on external datasets for better image feature extraction. In contrast, our model was trained without 2D label or additional dataset. With the joint detection and tracking paradigm, we do not need to reload or crop sensor data. Thus, the total time cost of our method is much less than other fusion-based tracking-by-detection methods.

\section{CONCLUSION}\label{sec:conclusion}
In this paper, we have presented an end-to-end camera-LiDAR fusion based joint multi-object detection and tracking system. Our model uses 2D and 3D paired data frames and produces 3D bounding boxes and association confidences for online mixed-integer programming. The proposed robust affinity computation and data association methods can greatly improve multi-object tracking performance. Without using additional training datasets, our method shows the state-of-the-art performance among camera-LiDAR fusion based MOT methods on the KITTI benchmark in terms of both MOTA (86.3\%) and processing speed (0.01s). Due to the fusion of camera and LiDAR data and the merge of object detection and tracking, our method is highly suitable for autonomous driving applications which demand high tracking robustness and real time performance.

\bibliographystyle{IEEEtran}
\bibliography{IEEEabrv,reference}

\clearpage

\addtolength{\textheight}{-12cm}   % This command serves to balance the column lengths

\end{document}